\documentclass[11pt,a4paper]{article}

\pdfoutput=1

\usepackage[hyperref]{acl2019}

\hypersetup{
    colorlinks=true,
    linkcolor=black,
    citecolor=black,
    filecolor=black,
    urlcolor=black,
}

\usepackage{times}
\usepackage{latexsym}
\usepackage{url}

\aclfinalcopy % Uncomment this line for the final submission
\def\aclpaperid{302} %  Enter the acl Paper ID here

\usepackage{amsmath,amsfonts,amssymb}
\usepackage{bm}
\usepackage{url}
\usepackage{xspace}
\usepackage{bbm}
\usepackage{booktabs}
\usepackage{enumitem}
\usepackage{comment}
\usepackage{standalone}
\usepackage{color}
\usepackage{pgfplots}
\usepackage{tikz}
\usetikzlibrary{fit}
\usetikzlibrary{calc}
\usepackage{colortbl}
\usepackage{pifont}% http://ctan.org/pkg/pifont
\usepackage{graphicx}
\graphicspath{figs}
\usepackage{filecontents}

\usetikzlibrary{arrows,decorations.markings}

\newcommand{\xmark}{\ding{55}}%

\renewcommand{\vec}[1]{\ensuremath \bm{#1}}
\newcommand{\mat}[1]{\ensuremath \bm{#1}}

\newcommand{\vk}{\ensuremath \vec{k}}

\newcommand{\vv}{\ensuremath \vec{v}}

\newcommand{\vx}{\ensuremath \vec{x}}
\newcommand{\vh}{\ensuremath \vec{h}}

\newcommand{\vphi}{\ensuremath \vec{\phi}}
\newcommand{\reals}{\mathbb{R}}

\newcommand{\system}[2][]{\texttt{#2#1}\xspace} %didn't know about xspace, that's cool -J
\newcommand{\gap}{\system{GAP}}

\newcommand{\example}[1]{\textit{#1}}
\newcommand{\mem}[1]{\ensuremath \texttt{m#1}}
\newcommand{\mzero}[0]{\mem{0}}
\newcommand{\mone}[0]{\mem{1}}

\title{The Referential Reader:\\A Recurrent Entity Network for Anaphora Resolution}

\author{Fei Liu \thanks{{} {} Work carried out as an intern at Facebook AI Research}\\
  The University of Melbourne\\
  Victoria, Australia
  \And
  Luke Zettlemoyer\\
  Facebook AI Research\\
  University of Washington\\
  Seattle, USA
  \And
  Jacob Eisenstein\\
  Facebook AI Research\\
  Seattle, USA}

\date{}

\begin{document}
\maketitle

\begin{abstract}
  We present a new architecture for storing and accessing entity mentions during online text processing.
  While reading the text, entity references are identified, and may be stored by either updating or overwriting a cell in a fixed-length memory.
  The update operation implies coreference with the other mentions that are stored in the same cell; the overwrite operation causes these mentions to be forgotten.
  By encoding the memory operations as differentiable gates, it is possible to train the model end-to-end, using both a supervised anaphora resolution objective as well as a supplementary language modeling objective.
  Evaluation on a dataset of pronoun-name anaphora demonstrates strong performance with purely incremental text processing.
\end{abstract}

\section{Introduction}
Reference resolution is fundamental to language understanding.
Current state-of-the-art systems employ the \emph{mention-pair} model, in which a classifier is applied to all pairs of spans~\cite[e.g.,][]{Lee+:2017}.
This approach is expensive in both computation and labeled data, and it is also cognitively implausible: human readers interpret text in a nearly online fashion~\cite{tanenhaus1995integration}.

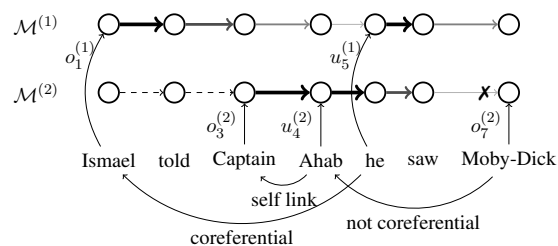
\begin{figure}[t]
    \begin{center}
        \resizebox{\linewidth}{!}{
            \begin{tikzpicture}
\tikzset{
  compnode/.style={draw,minimum size=1em,line width=1pt,circle}
}
  \matrix[column sep=1ex, row sep=4ex]{
    \node (m10) {$\mathcal{M}^{(1)}$};
    & \node[compnode] (m11) {};
    & \node[compnode] (m13) {};
    & \node[compnode] (m14) {};
    & \node[compnode] (m15) {};
    & \node[compnode] (m16) {};
    & \node[compnode] (m18) {};
    & \node[compnode] (m19) {};
    \\
    \node (m20) {$\mathcal{M}^{(2)}$};
    & \node[compnode] (m21) {};
    & \node[compnode] (m23) {};
    & \node[compnode] (m24) {};
    & \node[compnode] (m25) {};
    & \node[compnode] (m26) {};
    & \node[compnode] (m28) {};
    & \node[compnode] (m29) {};
    \\
    \node (x0) {};
    & \node (x1) {Ismael};
    & \node (x3) {told};
    & \node (x4) {Captain};
    & \node (x5) {Ahab};
    & \node (x6) {he};
    & \node (x8) {saw};
    & \node (x9) {Moby-Dick};
    \\
  };

  \path[->,bend left,at end,anchor=north east]
  (x1) edge node {$o_{1}^{(1)}$} (m11)
  (x6) edge node {$u_{5}^{(1)}$} (m16)
  ;

  \path[->,at end,anchor=north east]
  (x4) edge node {$o_{3}^{(2)}$} (m24)
  (x5) edge node {$u_{4}^{(2)}$} (m25)
  (x9) edge node {$o_{7}^{(2)}$} (m29)
  ;

  \path[->,bend left = 45,anchor=north]
  (x5) edge node {self link} (x4)
  (x6) edge node {coreferential} (x1)
  (x9) edge node {not coreferential} (x5)
  ;

  \path[->,line width=2.00pt,black!100] (m11) edge (m13);
  \path[->,line width=1.50pt,black!70]  (m13) edge (m14);
  \path[->,line width=1.00pt,black!50]  (m14) edge (m15);
  \path[->,line width=0.50pt,black!30]  (m15) edge (m16);
  \path[->,line width=2.00pt,black!100] (m16) edge (m18);
  \path[->,line width=1.00pt,black!50]  (m18) edge (m19);

  \path[->,dashed] (m21) edge (m23);
  \path[->,dashed] (m23) edge (m24);
  \path[->,line width=2.00pt,black!100] (m24) edge (m25);
  \path[->,line width=2.00pt,black!100] (m25) edge (m26);
  \path[->,line width=1.50pt,black!70]  (m26) edge (m28);
  \path[->,line width=0.50pt,black!30,at end,anchor=east] (m28) edge node {\textcolor{black!100}{\xmark}} (m29);

\end{tikzpicture}
        }
    \end{center}
    \caption{A referential reader with two memory cells. Overwrite and update are indicated by $o_{t}^{(i)}$ and $u_{t}^{(i)}$; in practice, these operations are continuous gates. Thickness and color intensity of edges between memory cells at neighboring steps indicate memory salience; \xmark{} indicates an overwrite.}    
    \label{fig:modelexample-narrow}
\end{figure}

We present a new method for reference resolution, which reads the text left-to-right while storing entities in a fixed-size working memory (\autoref{fig:modelexample-narrow}).
As each token is encountered, the reader must decide whether to: (a) link the token to an existing memory, thereby creating a coreference link, (b) overwrite an existing memory and store a new entity, or (c) disregard the token and move ahead.
As memories are reused, their salience increases, making them less likely to be overwritten.

This online model for coreference resolution is based on the memory network architecture~\cite{Weston+:2015},
in which memory operations are differentiable, enabling end-to-end training from gold anaphora resolution data.
Furthermore, the memory can be combined with a recurrent hidden state, enabling prediction of the next word.
This makes it possible to train the model from unlabeled data using a language modeling objective.

To summarize, we present a model that processes the text incrementally, resolving references on the fly~\cite{schlangen2009incremental}.
The model yields promising results on the \gap dataset of pronoun-name references.\footnote{Code available at: \url{https://github.com/liufly/refreader}}

\section{Model}

For a given document consisting of a sequence of tokens $\{w_t\}_{t=1}^T$, we represent text at two levels:
\setlist{nolistsep}
\begin{itemize}[noitemsep]
    \item Tokens: represented as $\{ \vx_t \}_{t=1}^T$, where the vector $\vx_t \in \reals^{D_x}$ is computed from any token-level encoder.%, and $T$ is the total number of tokens. 
    \item Entities: represented by a fixed-length memory $\mathcal{M}_t = \{(\vk^{(i)}_{t}, \vv^{(i)}_{t}, s^{(i)}_{t})\}_{i=1}^N$, where each memory is a tuple of a key $\vk^{(i)}_{t} \in \reals^{D_k}$, a value $\vv^{(i)}_{t} \in \reals^{D_v}$, and a salience $s^{(i)}_{t} \in [0,1]$.
\end{itemize}

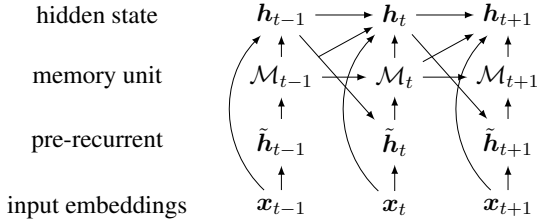
\begin{figure}[t]
    \begin{center}
        \resizebox{\columnwidth}{!}{
            \begin{tikzpicture}
\tikzset{
  compnode/.style={draw,minimum size=1em,line width=1pt,circle}
}
\pgfsetarrowsend{latex} 
\matrix[column sep=4ex, row sep=2ex]{
    \node {hidden state};
    & \node (h0) {$\vh_{t-1}$};
    & \node (h1) {$\vh_{t}$};
    & \node (h2) {$\vh_{t+1}$};
    \\
    \node {memory unit};
    & \node (m0) {$\mathcal{M}_{t-1}$};
    & \node (m1) {$\mathcal{M}_{t}$};
    & \node (m2) {$\mathcal{M}_{t+1}$};
    \\
    \node {pre-recurrent};
    & \node (ht0) {$\tilde{\vh}_{t-1}$};
    & \node (ht1) {$\tilde{\vh}_{t}$};
    & \node (ht2) {$\tilde{\vh}_{t+1}$};
    \\
    \node {input embeddings};
    & \node (x0) {$\vx_{t-1}$};
    & \node (x1) {$\vx_{t}$};
    & \node (x2) {$\vx_{t+1}$};
    \\
  };

  \path
  (x0) edge (ht0)
  (ht0) edge (m0)
  (m0) edge (h0)
  (m0) edge (m1)
  (m0) edge (h1)
  (h0) edge (h1)
  (h0) edge (ht1)
  (x1) edge (ht1)
  (ht1) edge (m1)
  (m1) edge (h1)
  (m1) edge (m2)
  (m1) edge (h2)
  (h1) edge (h2)
  (h1) edge (ht2)
  (x2) edge (ht2)
  (ht2) edge (m2)
  (m2) edge (h2)
  ;
  
  % \path[->,bend left = 45,scale=10]
%  \path[bend left=45,.tip={stealth[scale=500]}]%,StealthFill/.tip={Stealth[line width=1pt, scale=1.5]}, arrows={[round]}]]
  \path[bend left=45]%,StealthFill/.tip={Stealth[line width=1pt, scale=1.5]}, 
  (x0) edge (h0)
  (x1) edge (h1)
  (x2) edge (h2)
  ;

\end{tikzpicture}
        }
    \end{center}
    \caption{Overview of the model architecture.}
    \label{fig:modeloverview}
\end{figure}

There are two components to the model: the memory unit, which stores and tracks the states of the entities in the text; and the recurrent unit, which controls the memory via a set of gates. An overview is presented in \autoref{fig:modeloverview}.

\subsection{Recurrent Unit}
\label{sec:model-recurrent}
The recurrent unit is inspired by the Coreferential-GRU, in which the current hidden state of a gated recurrent unit~\cite[GRU;][]{Chung+:2014} is combined with the state at the time of the most recent mention of the current entity~\cite{Dhingra+:2018}. 
However, instead of relying on the coreferential structure to construct a dynamic computational graph, we use an external memory unit to keep track of previously mentioned entities and let the model learn to decide what to store in each cell.

The memory state is summarized by the weighted sum over values: $\vec{m}_t = \sum^N_{i=1} s^{(i)} \vv^{(i)}_t$.
The current hidden state and the input are combined into a \textbf{pre-recurrent state} $\tilde{\vh}_t = \tanh(\mat{W}\vh_{t-1} + \mat{U}\vx_t)$,
which is used to control the memory operations; the matrices $\mat{W}$ and $\mat{U}$ are trained parameters.
To compute the next hidden state $\vh_t$, we perform a recurrent update:
\begin{equation}
  \vh_{t} = \text{GRU}(\vx_t, (1 - c_t) \times \vh_{t-1} + c_t \times \vec{m}_t)
\end{equation}
where ${c_t= \min(\sigma(\mat{W}_c\tilde{\vh}_t + b_c), \sum_i s_t^{(i)})}$ is a gate that measures the importance of the memory network to the current token.
This gate is a sigmoid function of the pre-recurrent state, clipped by the sum of memory saliences.
This ensures that the memory network is used only when at least some memories are salient.

\subsection{Memory Unit}

The memory gates are a collection of scalars $\{(u_{t}^{(i)}, o_{t}^{(i)})\}_{i=1}^N$, indicating \textbf{u}pdates and \textbf{o}verwrites to cell $i$ at token $w_t$.
To compute these gates, we first determine whether $w_t$ is an entity mention, using a sigmoid-activated gate
${e_t = \sigma(\vphi_e \cdot \tilde{\vh}_t)},$
where $\vphi_e \in \reals^{D_h}$ is a learnable vector.
We next decide whether $w_t$ refers to a \emph{previously} mentioned entity:
${r_t = \sigma(\vphi_r \cdot \tilde{\vh}_t) \times e_t},$
where $\vphi_r \in \reals^{D_h}$ is a learnable vector. 

\paragraph{Updating existing entities.}
If $w_t$ is a referential entity mention (${r_t \approx 1}$), it may refer to an entity in the memory.
To compute the compatibility between $w_t$ and each memory, we first summarize the current state as a query vector,
$\vec{q}_t = f_q(\tilde{\vh}_t),$
where $f_q$ is a two-layer feed-forward network.
The query vector is then combined with the memory keys and the reference gate to obtain attention scores,
${\alpha_{t}^{(i)} = r_t \times \text{SoftMax}(\vk^{(i)}_{t-1} \cdot \vec{q}_t + b)},$
where the softmax is computed over all cells $i$, and $b$ is a learnable bias term, inversely proportional to the likelihood of introducing a new entity. 
The update gate is then set equal to the query match $\alpha_{t}^{(i)}$, clipped by the salience,
$u_{t}^{(i)} = \min(\alpha_{t}^{(i)}, 2 s_{t-1}^{(i)}).$
The upper bound of $2s_{t-1}^{(i)}$ ensures that an update can at most triple the salience of a memory.

\paragraph{Storing new entities.}
Overwrite operations are used to store new entities. 
The total amount to overwrite is ${\tilde{o}_{t} = e_t - \sum^N_{i=1} u_{t}^{(i)}}$, which is the difference between the entity gate and the sum of the update gates.
We prefer to overwrite the memory with the lowest salience.
This decision is made differentiable using the Gumbel-softmax distribution~\cite[GSM;][]{Jang+:2017}, ${o_t^{(i)} = \tilde{o}_{t} \times \text{GSM}^{(i)}(-\vec{s}_{t-1}, \tau)}$ and ${\vec{s}_{t}=\{s^{(i)}_t\}_{i=1}^{N}}$.\footnote{Here $\tau$ is the ``temperature'' of the distribution, which is gradually decreased over the training period, until the distribution approaches a one-hot vector indicating the argmax.}

\paragraph{Memory salience.}
To the extent that each memory is not updated or overwritten, it is copied along to the next timestep. 
The weight of this copy operation is:
${r_{t}^{(i)} = 1 - u_{t}^{(i)} - o_{t}^{(i)}}.$
The salience decays exponentially,
\begin{align}
  \lambda_t = &{} (e_t \times \gamma_e + (1 - e_t) \times \gamma_n)\\
  s^{(i)}_t = &{} \lambda_t \times r_{t}^{(i)} \times s^{(i)}_{t-1} + u_{t}^{(i)} + o_{t}^{(i)},
\end{align}
where $\gamma_e$ and $\gamma_n$ represent the salience decay rate upon seeing an entity or non-entity.\footnote{We set $\gamma_{e} = \exp(\log(0.5)/\ell_{e})$ with $\ell_e = 4$ denoting the entity half-life, which is the number of entity mentions before the salience decreases by half. The non-entity halflife $\gamma_n$ is computed analogously, with $\ell_n = 30$.}

\paragraph{Memory state.}
To update the memory states, we first transform the pre-recurrent state $\tilde{\vh}_t$ into the memory domain, obtaining overwrite candidates for the keys and values,
$\tilde{\vec{k}}_t = f_{k}(\tilde{\vh}_t)$ and $\tilde{\vec{v}}_t = f_{v}(\tilde{\vh}_t),$
where $f_{k}$ is a two-layer residual network with $\tanh$ nonlinearities, and $f_{v}$ is a linear projection with a $\tanh$ non-linearity.
Update candidates are computed by GRU recurrence with the overwrite candidate as the input.
This yields the state update,
\begin{align*}
    \vk^{(i)}_t &= u_{t}^{(i)}\text{GRU}_k(\vk^{(i)}_{t-1}, \tilde{\vec{k}}_t) + o_{t}^{(i)}\tilde{\vec{k}}_t + r_{t}^{(i)} \vk^{(i)}_{t-1}\\
    \vv^{(i)}_t &= u_{t}^{(i)}\text{GRU}_v(\vv^{(i)}_{t-1}, \tilde{\vec{v}}_t) + o_{t}^{(i)}\tilde{\vec{v}}_t + r_{t}^{(i)} \vv^{(i)}_{t-1}.
\end{align*}

\subsection{Coreference Chains}
\label{sec:coref-chains}
To compute the probability of coreference between the mentions $w_{t_1}$ and $w_{t_2}$, we first compute the probability that each cell $i$ refers to the same entity
at both of those times, 
\begin{equation}
  \omega_{t_1,t_2}^{(i)} = \prod_{t = t_1+1}^{t_2} (1 - o_{t}^{(i)})
\end{equation}
Furthermore, the probability that mention $t_1$ is stored in memory $i$ is ${u_{t_1}^{(i)} + o_{t_1}^{(i)}}$. 
The probability that two mentions corefer is then the sum over memory cells,
\begin{align}
\hat{\psi}_{t_1,t_2} &= \sum_{i=1}^N (u_{t_1}^{(i)} + o_{t_1}^{(i)}) \times u_{t_2}^{(i)} \times \omega_{t_1, t_2}^{(i)}.
\label{eq:coref}
\end{align}

\subsection{Training}
\label{sec:training}
The coreference probability defined in \autoref{eq:coref} is a differentiable function of the gates, which in turn are computed from the inputs ${w_1, w_2, \ldots w_T}$.
We can therefore train the entire network end-to-end from a cross-entropy objective, where a loss is incurred for incorrect decisions on the level of token pairs.
Specifically, we set $y_{i,j} = 1$ when $w_i$ and $w_j$ corefer (coreferential links), and also when both $w_i$ and $w_j$ are part of the same mention span (self links).
The coreference loss is then the cross-entropy $\sum_{i=1}^T \sum_{j=i+1}^T H(\hat{\psi}_{i,j}, y_{i,j})$.

Because the hidden state $\vh_t$ is computed recurrently from $w_{1:t}$, the reader can also be trained from a language modeling objective, even when coreference annotations are unavailable.
Word probabilities $P(w_{t+1} \mid \vh_t)$ are computed by projecting the hidden state $\vh_t$ by a matrix of output embeddings, and applying the softmax operation.

\section{Experiments}
As an evaluation of the ability of the referential reader to correctly track entity references in text, we evaluate against the \gap{} dataset, recently introduced by \newcite{Webster+:2018}.
Each instance consists of: (1) a sequence of tokens $w_1,\ldots,w_T$ extracted from Wikipedia biographical pages; (2) two person names ($A$ and $B$, whose token index spans are denoted $s_A$ and $s_B$); (3) a single-token pronoun ($P$ with the token index $s_P$); and (4) two binary labels ($y_A$ and $y_B$) indicating whether $P$ is referring to $A$ or $B$.

\paragraph{Language modeling.}
Given the limited size of \gap, it is difficult to learn a strong recurrent model.
We therefore consider the task of language modeling as a pre-training step.
We make use of the page text of the original Wikipedia articles from \gap, the URLs to which are included as part of the data release.
% \footnote{Page text downloaded with the Python Wikipedia API v1.4.0: \url{https://pypi.org/project/wikipedia/}.}
This results in a corpus of $3.8$ million tokens, which is used for pre-training.
The reader is free to use the memory to improve its language modeling performance, but it receives no supervision on the coreference links that might be imputed on this unlabeled data.

\paragraph{Prediction.}
At test time, we make coreference predictions using the procedure defined in \autoref{sec:coref-chains}.
Following \newcite{Webster+:2018}, we do not require exact string match for mention detection: if the selected candidate is a substring of the gold span, we consider it as a predicted coreferential link between the pronoun and person name.
Concretely, we focus on the token index $s_P$ of the pronoun and predict the positive coreferential relation of the pronoun $P$ and person name $A$ if any (in the span of $s_A$) of $\hat{\psi}_{s_A,s_P}$ (if $s_A < s_P$) or $\hat{\psi}_{s_P,s_A}$ (otherwise) is greater than a threshold value (selected on the validation set).\footnote{As required by \newcite{Webster+:2018}, the model is responsible for detecting mentions; only the scoring function accesses labeled spans.}

\paragraph{Evaluation.}
Performance is measured on the \gap{} test set, using the official evaluation script.
We report the overall $F_1$, as well as the scores by gender (\textbf{M}asculine: $F_1^M$ and \textbf{F}eminine: $F_1^F$), and the bias (the ratio of $F_1^F$ to $F_1^M$: $\frac{F_1^F}{F_1^M}$).

\paragraph{Systems.}
We benchmark our model (RefReader) against a collection of strong baselines presented in the work of \newcite{Webster+:2018}:
(1) a state-of-the-art mention-pair coreference resolution~\cite{Lee+:2017};
(2) a version of (1) that is retrained on \gap{};
(3) a rule-based system based on syntactic parallelism~\cite{Webster+:2018};
(4) a domain-specific variant of (3) that incorporates the lexical overlap between each candidate and the title of the original Wikipedia page~\cite{Webster+:2018}.
We evaluate a configuration of RefReader that uses two memory cells; other details are in the supplement (\autoref{app:config}).

\paragraph{Results.}

\begin{table}[tb]
    \resizebox{\columnwidth}{!}{%
              \begin{tabular}{lcccc}
            \toprule
                                         & $F_1^M$ & $F_1^F$ & $\frac{F_1^F}{F_1^M}$ & $F_1$ \\
            \midrule
            %\newcite{Lee+:2013}$\dagger$ & 53.4 & 47.5 & 0.89 & 50.5 \\
            \newcite{Clark+:2015}$\dagger$ & 53.9 & 52.8 & 0.98 & 53.3 \\
            %\newcite{Wiseman+:2016}$\dagger$ & 67.8 & 59.1 & 0.87 & 63.6 \\
            \newcite{Lee+:2017}$\dagger$ & 67.7  & 60.0  & 0.89 & 64.0  \\
            \newcite{Lee+:2017}, re-trained & 67.8 & 66.3 & 0.98 & 67.0 \\[1ex]
            Parallelism$\dagger$         & 69.4  & 64.4  & 0.93 & 66.9  \\
            Parallelism+URL$\dagger$     & 72.3  & 68.8  & 0.95 & 70.6  \\[1ex]
            RefReader, LM objective$\ddagger$               & 61.6  & 60.5  & 0.98 & 61.1  \\
            RefReader, coref objective$\ddagger$            & 69.6  & 68.1  & 0.98 & 68.9  \\
            RefReader, LM + coref$\ddagger$       & \textbf{72.8}  & \textbf{71.4}  & \textbf{0.98} & \textbf{72.1}  \\[1ex]
            RefReader, coref + BERT$\star$ & \textbf{80.3} & \textbf{77.4} & \textbf{0.96} & \textbf{78.8}\\
            \bottomrule
        \end{tabular}

    }
    \caption{\gap test set performance. $\dagger$: reported in \newcite{Webster+:2018}; $\ddagger$: strictly incremental processing; $\star$: average over $5$ runs with different random seeds.}
    \label{tbl:performance}
\end{table}

As shown in \autoref{tbl:performance}, RefReader achieves state-of-the-art performance,
outperforming strong pretrained and retrained systems~\cite[e.g.,][]{Lee+:2017}, as well as domain-specific heuristics (Parellelism+URL). 
Language model pretraining yields an absolute gain of $3.2$ in $F_1$.
This demonstrates the ability of RefReader to leverage unlabeled text, which is a distinctive feature in comparison with prior work.
When training is carried out in the unsupervised setting (with the language modeling objective only), the model is still capable of learning the latent coreferential structure between pronouns and names to some extent, outperforming a supervised coreference system that gives competitive results on OntoNotes~\cite{Clark+:2015}.

We also test a combination of RefReader and BERT~\cite{devlin2018bert}, using BERT's contextualized word embeddings as base features $\vx_t$ (concatenation of the top 4 layers), which yields substantial improvements in accuracy.
While this model still resolves references incrementally, it cannot be said to be \emph{purely} incremental, because BERT uses ``future'' information to build its contextualized embeddings.\footnote{Future work may explore the combination of RefReader and large-scale pretrained incremental language models~\cite[e.g.,][]{radford2019language}.}
Note that the gender bias increases slightly, possibly due to bias in the data used to train BERT.

\gap examples are short, containing just a few entity mentions. 
To test the applicability of our method to longer instances, we produce an alternative test set in which pairs of \gap instances are concatenated together, doubling the average number of tokens and entity mentions. 
Even with a memory size of two, performance drops to $F_1 = 70.2$ (from $72.1$ on the original test set). 
This demonstrates that the model is capable of reusing memory cells when the number of entities is larger than the size of the memory. 
We also test a configuration of RefReader with four memory cells, and observe that performance on the original test set decreases only slightly, to $F_1 = 71.4$ (against RefReader LM + coref).

\begin{figure*}[tb]
  \centering
  \resizebox{.99\textwidth}{!}{
    \includegraphics[width=\textwidth]{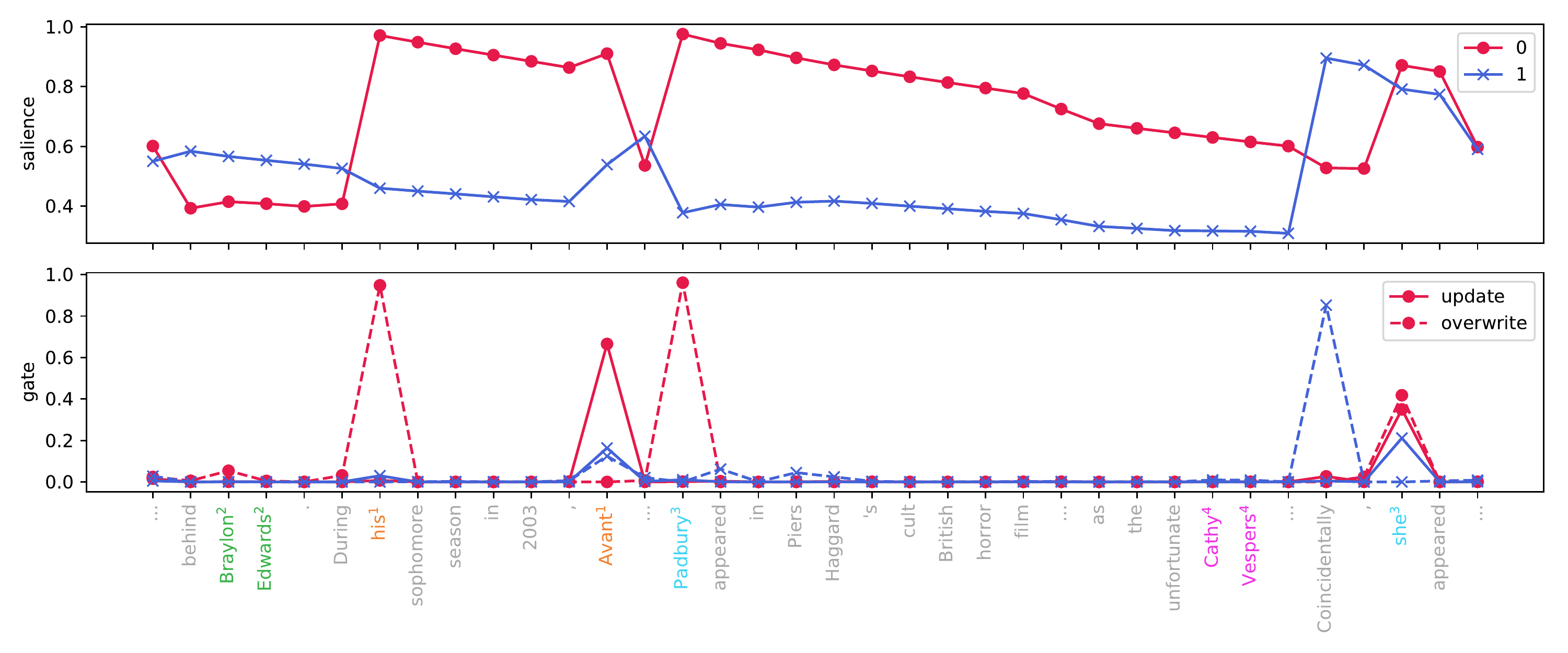}
  }
  \caption{An example of the application the referential reader to a concatenation of two instances from \gap. The ground truth is indicated by the color of each token on the $x$-axis as well as the superscript.}
  \label{fig:example}
\end{figure*}

\paragraph{Case study and visualization.}
\autoref{fig:example} gives an example of the behavior of the referential reader, as applied to a concatenation of two instances from \gap.\footnote{For an example involving multi-token spans, see \autoref{app:examples}.} The top panel shows the salience of each entity as each token is consumed, with the two memory cells distinguished by color and marker. The figure elides long spans of tokens whose gate activations are nearly zero. These tokens are indicated in the $x$-axis by ellipsis; the corresponding decrease in salience is larger, because it represents a longer span of text. The bottom panel shows the gate activations for each token, with memory cells again distinguished by color and marker, and operations distinguished by line style. The gold token-entity assignments are indicated with color and superscript.

The reader essentially ignores the first name, \example{Braylon Edwards}, making a very weak overwrite to memory 0 ($\mzero$). It then makes a large overwrite to $\mzero$ on the pronoun \example{his}. When encountering the token \example{Avant}, the reader makes an update to the same memory cell, creating a cataphoric link between \example{Avant} and \example{his}. The name \example{Padbury} appears much later (as indicated by the ellipsis), and at this point, $\mzero$ has lower salience than $\mone$. For this reason, the reader chooses to overwrite $\mzero$ with this name. The reader ignores the name \example{Cathy Vespers} and overwrites $\mone$ with the adverb \example{coincidentally}. On encountering the final pronoun \example{she}, the reader is conflicted, and makes a partial overwrite to $\mzero$, a partial update (indicating coreference with \example{Padbury}), and a weaker update to $\mone$. If the update to $\mzero$ is above the threshold, then the reader may receive credit for this coreference edge, which would otherwise be scored as a false negative.

The reader ignores the names \example{Braylon Edwards}, \example{Piers Haggard}, and \example{Cathy Vespers}, leaving them out of the memory. \example{Edwards} and \example{Vespers} appear in prepositional phrases, while \example{Haggard} is a possessive determiner of the object of a prepositional phrase. Centering theory argues that these syntactic positions have low salience in comparison with subject and object position~\cite{grosz1995centering}. It is possible that the reader has learned this principle, and that this is why it chooses not to store these names in memory. However, the reader also learns from the \gap supervision that pronouns are important, and therefore stores the pronoun \example{his} even though it is also a possessive determiner.
\section{Related Work}
Memory networks provide a general architecture for online updates to a set of distinct memories~\cite{Weston+:2015,Sukhbaatar+:2015}. The link between memory networks and incremental text processing was emphasized by \newcite{cheng-dong-lapata:2016:EMNLP2016}. \newcite{Henaff+:2017} used memories to track the states of multiple entities in a text, but they predefined the alignment of entities to memories, rather than learning to align entities with memories using gates. The incorporation of entities into language models has also been explored in prior work~\cite{Yang+:2017,Kobayashi+:2017}; similarly, \newcite{Dhingra+:2018} augment the gated recurrent unit (GRU) architecture with additional edges between coreferent mentions. In general, this line of prior work assumes that coreference information is available at test time (e.g., from a coreference resolution system), rather than determining coreference in an online fashion. \newcite{Ji+:2017} propose a generative entity-aware language model that incorporates coreference as a discrete latent variable. For this reason, importance sampling is required for inference, and the model cannot be trained on unlabeled data.

\section{Conclusion}
This paper demonstrates the viability of incremental reference resolution, using an end-to-end differentiable memory network.
This enables semi-supervised learning from a language modeling objective, which substantially improves performance.
A key question for future work is the performance on longer texts, such as the full-length news articles encountered in OntoNotes.
Another direction is to further explore semi-supervised learning, by reducing the amount of training data and incorporating linguistically-motivated constraints based on morphosyntactic features.

\section*{Acknowledgments}

We would like to thank the anonymous reviewers for their valuable feedback, Yinhan Liu, Abdel-rahman Mohamed, Omer Levy, Kellie Webster, Vera Axelrod, Mandar Joshi, Trevor Cohn and Timothy Baldwin for their help and comments.

\bibliography{acl2019}
\bibliographystyle{acl_natbib}

\appendix
\section{Supplemental information}
\label{app:config}
\paragraph{Model configuration.}
Training is carried out on the development set of \system{GAP} with the Adam optimizer~\cite{Kingma+:2014} and a learning rate of $0.001$.
Early stopping is applied based on the performance on the validation set. 
We use the following hyperparameters:
\begin{itemize}
\item embedding size $D_x=300$;
\item memory key size $D_k=16$ ($32$ with BERT) and value size $D_v=300$; the hidden layers in the memory key/value updates $f_k$ and $f_v$ are also set to $16$ ($32$ with BERT) and $300$ respectively;
\item number of memory cells $N=2$;
\item pre-recurrent and hidden state sizes $D_h=300$;
\item salience half-life for words and entity mentions are $30$ and $4$ respectively;
\item Gumbel softmax starts at temperature $\tau = 1.0$ with an exponential decay rate of $0.5$ applied every $10$ epochs;
\item dropout is applied to the embedding layer, the pre-recurrent state $\tilde{\vh}_t$, and the GRU hidden state $\vh_t$, with a rate of $0.5$;
\item self and coreferential links are weighted differently in the coreference loss cross-entropy in \autoref{sec:training} with $0.1$ and $5.0$ and negative coreferential links weighted higher than positive ones with a ratio of 10:1 to penalize false positive predictions.
\end{itemize}
For the RefReader model trained only on coreference annotations, the base word embeddings ($\vx_t$) are fixed to the pretrained GloVe embeddings~\cite{Pennington+:2014}.
In the RefReader models that include language model pretraining, embeddings are learned on the language modeling task.
Language modeling pre-training is carried out using the same configuration as above; the embedding update and early stopping are based on perplexity on a validation set.

\section{Multi-token Span Example}
\label{app:examples}

\begin{figure*}[tb]
  \centering
  \includegraphics[width=\textwidth]{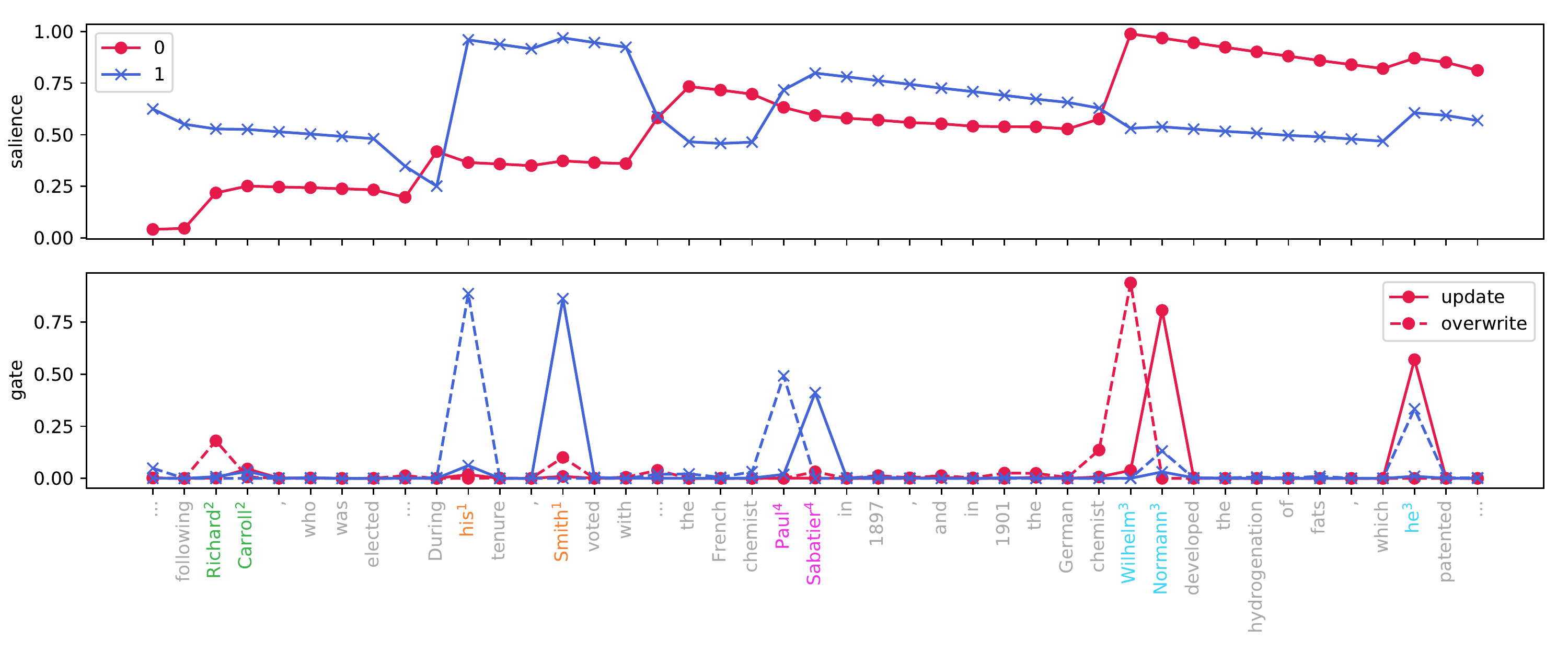}
  \caption{Another example of the referential reader, as applied to a concatenation of two instances from \gap. Again, the ground truth is indicated by the color of each token on the $x$-axis as well as the superscript.}
  \label{fig:example-2}
\end{figure*}

In the example shown in \autoref{fig:example-2}, the system must handle multi-token spans \example{Paul Sabatier} and \example{Wilhelm Normann}. It does this by overwriting on the first token, and updating on the second token, indicating that both tokens are part of the name of a single entity. The reader also correctly handles an example of cataphora (\example{During \textbf{his} tenure, \textbf{Smith} voted \ldots}). It stores \example{Paul Sabatier} in the same memory as \example{Smith}, but overwrites that memory so as not to create a coreference link. The reader reuses memory one for both entities because in the intervening text, memory zero acquired more salience. Finally, the model perceives some ambiguity on the pronoun \example{he} at the end: it narrowly favors coreference with \example{Normann}, but assigns some probability to the creation of a new entity.

\end{document}